# Evaluating the Portability of an NLP System for Processing Echocardiograms: A Retrospective, Multi-site Observational Study

Prakash Adekkanattu, PhD[1], Guoqian Jiang, MD PhD[2], Yuan Luo, PhD[3], Paul R. Kingsbury, PhD[2], Zhenxing Xu, PhD[1], Luke V. Rasmussen, MS[3], Jennifer A. Pacheco, MS[3], Richard C. Kiefer, MS[2], Daniel J. Stone, MS[2], Pascal S. Brandt, MS[4], Liang Yao, PhD[3], Yizhen Zhong, BS[3], Yu Deng, BS[3], Fei Wang, PhD[1], Jessica S. Ancker, MPH PhD[1], Thomas R. Campion, Jr, PhD[1], Jyotishman Pathak, PhD[1]

[1]Weill Cornell Medicine, New York, NY, [2]Mayo Clinic, Rochester, MN, [3]Northwestern University, Chicago, IL, [4]University of Washington, Seattle, WA

## Abstract

*While natural language processing (NLP) of unstructured clinical narratives holds the potential for patient care and clinical research, portability of NLP approaches across multiple sites remains a major challenge. This study investigated the portability of an NLP system developed initially at the Department of Veterans Affairs (VA) to extract 27 key cardiac concepts from free-text or semi-structured echocardiograms from three academic medical centers: Weill Cornell Medicine, Mayo Clinic and Northwestern Medicine. While the NLP system showed high precision and recall measurements for four target concepts (aortic valve regurgitation, left atrium size at end systole, mitral valve regurgitation, tricuspid valve regurgitation) across all sites, we found moderate or poor results for the remaining concepts and the NLP system performance varied between individual sites.*

## Introduction

Echocardiography is currently the most widely used non-invasive cardiac imaging in the diagnosis, treatment, and follow-up of patients with cardiovascular diseases. Various measurements in echocardiography provide key insights into the mechanisms and health of a patient's cardiovascular system[1]. Aortic, mitral and tricuspid valves are assessed for abnormal structure, thickness, and valve dysfunction (regurgitation, stenosis). Increased values for chamber sizes indicate possible dilation. The thicknesses of the interventricular septum and posterior left ventricular (LV) wall are used to determine the presence of hypertrophy. Left ventricular systolic performance indicates severity of heart disease, and most laboratories quantify Left Ventricular Ejection Fraction (LVEF) to a value ranging from 5-75 percent of total blood volume. Echocardiography measurements are generally stored in electronic health record (EHR) systems as an echocardiogram (echo) report, which is frequently an unstructured or semi-structured text document. This limits the downstream use of the report data by automated systems for clinical research (e.g., phenotyping), care management and point-of-care clinical decision support[2].

Natural language processing of unstructured clinical narratives, such as echo reports, holds tremendous potential to extract meaningful information and fill this knowledge gap[3-6]. NLP utilizes various algorithms to automatically extract relevant clinical information from free text and semi-structured data sources. However, a recent review of the literature by Demner-Fushman and Elhadad suggests that NLP remains an "emerging technology", with a significant gap between promise and reality[7]. One of the major challenges in broader adoption of NLP systems and technologies is their portability across multiple EHR systems. While several studies, including work done by our study team, have demonstrated varying levels of success in portability of NLP technologies across institutions[8-12], recent work by Carrell et al. argue that there remain significant challenges in adapting NLP systems across multiple sites, which include assembling clinical corpora, managing diverse document structures and handling idiosyncratic linguistic expressions[13]. These barriers and the requirement for significant upfront "investment" limit wide-scale adoption of NLP tools and systems across health systems, particularly those in low-resource settings and environments.

To understand these barriers further, in this study we investigated the implementation of an existing NLP system - Leo, developed by the Department of Veterans Affairs (VA) Informatics and Computing Infrastructure (VINCI) - across three academic medical centers. The goal of the current study was to assess the portability of *EchoExtractor* by installing, without any system modifications whatsoever, and evaluating its performance at three non-VA medical centers. This project, as part of the Phenotype Execution and Modeling Architecture (PhEMA), was motivated by the fact that EHR based computational phenotyping using both structured and unstructured data is a key requirement for deriving high quality phenotypes from EHRs[14], and a portable NLP system that can be run across multiple institutions is much needed[15,16].

**Materials and Methods**

System Description and Target Concepts

Leo provides a set of services and libraries that facilitate the rapid creation and deployment of NLP analysis tools. A major strength of this approach is the use of Unstructured Information Management Architecture (UIMA), which provides a set of standards for creating, discovering, composing and deploying a broad range of text analytics capabilities and integrating them with search technologies[17]. This standard is very attractive for NLP research because it is open-source (available through Apache[18]), and enables incorporation of NLP tools developed elsewhere. The distribution package for *EchoExtractor* was made available through a VA GitHub repository[19]. Installation of *EchoExtractor* using UIMA-AS 2.8.1 was deemed to be relatively straightforward, and was completed within a day at WCM and Northwestern. Installation at Mayo Clinic was performed using installation instructions found for Leo[20] where an alternate version of UIMA (UIMA-AS 2.9.0) was installed to support Leo, yet the installation process was still also completed within a day. We tested the performance of the pipeline without any custom modification to evaluate the true portability of the NLP system.

While previous studies have shown success in extracting echocardiography measurements from Echo reports[12] using NLP methods, the focus has been primarily on extracting one key concept - LVEF. The VINCI team created a specific instance of Leo named *EchoExtractor* to extract 27 different important cardiac concepts from echo and radiology reports[21]. A preliminary analysis of results obtained after processing 100 echo reports from WCM center through the NLP pipeline revealed that not all target concepts were present. Concepts such as left ventricular contractility, left ventricular hypertrophy and aortic valve max pressure gradient were not identified. Therefore we focussed on the remaining 24 elements in our in extended evaluation study. Later, we observed that the above three concepts were in fact present in echo reports originated from other centers, however we did not include them in our performance evaluation. The 24 target concepts we investigated in this study is listed in Table 1. For a description of these concepts, we refer readers to Patterson, et al.[21].

Table 1: List of target concepts extracted from echo reports at Weill Cornell Medicine, Mayo Clinic and Northwestern Medicine sites

| | |
|---|---|
| aortic valve mean gradient | left ventricular posterior wall thickness at end diastole |
| aortic valve orifice area | mitral valve mean gradient |
| aortic valve regurgitation | mitral valve orifice area |
| aortic valve regurgitation peak velocity | mitral valve regurgitation |
| aortic valve stenosis | mitral valve regurgitation peak velocity |
| e/e prime ratio | mitral valve stenosis |
| inter-ventricular septum dimension at end diastole | pulmonary artery pressure |
| left atrium size at end systole | right atrial pressure |
| left ventricular dimension at end diastole | tricuspid valve mean gradient |
| left ventricular dimension at end systole | tricuspid valve orifice area |
| left ventricular size | tricuspid valve regurgitation |
| left ventricular ejection fraction | tricuspid valve regurgitation peak velocity |

Data collection

We obtained Echo reports from all three medical centers participating in this study. Weill Cornell Medicine (WCM) uses the EpicCare® Ambulatory EHR platform to document clinical care, from which we collected all echocardiograms used as the basis of this study. We obtained 200 echocardiograms for 200 patients who were 18 years or older at the time of the echocardiogram results and one of the study team members (PA) manually annotated the data elements extracted by the *EchoExtractor* to create the gold standard at WCM. These reports in general formatted into multiple sections containing semi-structured, and unstructured data. Reports have a subject meta-data section, one or more tabular sections, and a comment section. The tabular sections contain quantitative measures such as wall thicknesses, chamber dimensions, or flow velocities. Unstructured fields contain descriptions of clinically relevant findings as interpreted by the technician. For example, the descriptions could say "Normal aortic valve and aortic root", "Normal left ventricular size and function", "There is trace mitral regurgitation". Furthermore, left ventricular function is often subjectively quantified as "mild", "moderate", "mildly to moderately reduced" or "severely reduced".

Mayo Clinics implemented an Unified Data Platform (UDP) to provide practical data solutions that creates a combined view of multiple heterogeneous EHR data sources (including Epic) through effective data orchestration, along with a number of data marts based on common data models. We retrieved the echocardiogram reports from the Mayo Clinic UDP platform from patients over 18 years old at the time of the echocardiogram results after verifying authorization to use the data in accordance with the Minnesota Health Records Act. Patient records were found in two formats: Rich Text Format (2000 to 2012), and Portable Document Format (PDF) from 2012 to 2018. For the echocardiograms generated between 2012-18, using a custom-built Python application, original text from the PDF documents was extracted. We randomly selected 200 echocardiogram reports for this study from each document format. The text portions of these reports were extracted, PHI was censored, and other elements such as visual elements (charts and graphics) and metadata were discarded. Echo reports found at Mayo Clinic generally consisted of three types of sections: unstructured text, numbered final impressions, and semi-structured tables of echocardiogram data. These reports were manually annotated with data elements extracted by the *EchoExtractor* by an experienced nurse at the Mayo Clinic which generated an annotation corpus as Mayo Clinic's gold standard.

At Northwestern Medicine, we retrieved 251,521 free-text echocardiogram reports for patients over 18 years old at the time of the echocardiogram between 2000 and 2017 from Northwestern Medicine® Enterprise Data Warehouse (NMEDW). The NMEDW hosts a comprehensive and integrated repository of the most used clinical and research data sources, at the time from 2 hospitals: Northwestern Memorial Hospital (NMH) and Lake Forest Hospital (LFH). Echo reports typically contained a mixture of semi-structured tables of echocardiogram data, along with unstructured impressions and findings. Overall structure of reports was primarily a metadata section containing patient and procedure information, semi-structured tabular results, followed by the impression (although many variations in this structure were observed). We randomly selected 200 echocardiogram reports and multiple study team members (YZ, YD, LY, JP, LR, YL) manually annotated the data elements extracted by the *EchoExtractor*. Across reviewers, we have discordant findings reconciled in a group setting to ensure consistent coding and interpretation, sometimes in consultation with a cardiovascular physician colleague. The final consolidated list of annotations was used as the gold standard at Northwestern.

We also obtained echocardiograms from the Medical Information Mart for Intensive Care III (MIMIC-III) database[22]. The latest version, MIMIC-III, contains de-identified patient records for >40,000 critical care patients admitted to Beth Israel Deaconess Medical Center between 2001 and 2012. We extracted 44,559 echo reports from 21,645 adult patients who were 18 years or older. We then randomly selected 200 echo reports for 200 unique patients and a study team member (PA) manually annotated the data elements extracted by the *EchoExtractor* to create the gold standard for the MIMIC dataset. Unlike the Echo reports from three medical centers, MIMIC reports were in free-text format with no tabular or semi-structured sections. In general, these reports have an interpretation section comprising subsections for findings and conclusions. Findings section has detailed descriptions about left atrium, right atrium, left ventricle, right ventricle, aorta, aortic valve, mitral valve, tricuspid valve, and pulmonic valve/pulmonic artery. MIMIC reports are for patients admitted to the intensive care unit; hence, patient characteristics are different from those patients treated at various outpatient settings in WCM, Mayo Clinic and Northwestern. At each medical center we limited the number of reports for manual review to 200 mainly due to limited resource availability, and also due to the fact that, in general, echo reports tend to follow same format(s) within each clinical setting.

Extraction methodology

Patterson et al. has described in detail the logic for concept extraction employed in the present study[21]. Various target concepts were extracted through a multi-stage process in a rule-based algorithm that involves identifying a concept mention in the text, identifying a value mention in the text, and linking appropriate concept and value mentions in a relationship. Term identification and term mapping was used to detect a concept mention in the text. Term identification was performed using regular expressions matching and term mapping was achieved by building a custom term lookup dictionary. Value identification was achieved through regular expressions that match both quantitative and qualitative values for concepts. Finally, a relationship between a concept and value was established through pattern matching, from a set of patterns identified for the clinical subdomain of echocardiograms.

Reference standard and performance evaluations

For all three study sites and for the MIMIC dataset, we determined performance of the NLP pipeline individually by developing independent gold standard datasets (see above). Results of *EchoExtractor* output were tabulated and compared against the gold standard through manual reviews. For each concept, a reviewer compared the output generated by *EchoExtractor* to each of the 200 reports. The reviewer identified all mentions of a concept and the associated quantitative or qualitative values. The reviewer also confirmed that the concepts not found by *EchoExtractor* in a given report were truly not mentioned in the respective report. The reviewer identified both numeric values as well as qualitative descriptions such as normal, mild, moderate, dilated, and severe. If a numeric value is documented as a range, then we identified both the values as the minimum and maximum. In reports containing multiple instances of a concept, we identified all instances. In cases where concept value is expressed using a greater than or less than symbol (e.g. LVEF >55), the reviewer simply identified the value ignoring the symbol. Some echo reports express uncertainty about certain concepts such as LVEF value using a question mark (e.g. LVEF ?55-70). In such cases, the reviewer extracted the value, ignoring the question mark.

Each report was classified as one of four possible cases: true positive (report had the concept present and *EchoExtractor* identified the "concept-value pair" and matched with the value given in the reference standard); false positive (report had the concept absent, but *EchoExtractor* produced a concept-value pair); true negative (report had the concept absent, and *EchoExtractor* did not find any concept-value pair); and false negative (report had the concept present, but *EchoExtractor* did not identify a concept-value pair, or the value extracted did not match with the corresponding value in the text). A value for a concept can be either a quantitative measure as in "systolic function (LVEF>55%)" for ejection fraction, or a qualitative assessment as in "no aortic regurgitation". Note that we are not assessing if a given concept is present or absent in the report, rather whether the NLP method correctly identified the concept-value pairs documented in the report. For example, for a statement such as "There is no aortic valve stenosis", if the NLP system extracted 'no' as the assessment for the concept 'aortic valve stenosis', we will consider this as a true positive case. For a given target concept, when there were multiple instances of concept-value pair extracted by *EchoExtractor*, we used the following heuristic measure to select a given instance of the concept in order to compare directly with the reference standard: we selected the last concept-value pair, normally in the conclusion part of the report (as the order of results and text in the reports was usually chronological). The total outcomes of the four possible cases were then used to calculate various statistical performance measures, namely, precision (positive predictive value), recall (sensitivity or true positive rate), and the F-score (the harmonic mean of recall and precision).

**Results**

Results of *EchoExtractor* validation performed on echo reports from WCM, Mayo, Northwestern and MIMIC datasets are presented in Table 2. Among all datasets aortic valve regurgitation, aortic valve stenosis, left ventricular size, LVEF, mitral valve regurgitation, pulmonary artery pressure, and tricuspid valve regurgitation were the most frequently mentioned concepts. Other concepts were mentioned infrequently or in some cases absent all together as indicated by the letter 'A' in Table 2. While target concepts such as aortic valve regurgitation peak velocity, mitral valve regurgitation peak velocity, tricuspid valve mean gradient, and tricuspid valve orifice area were absent from the echo reports from WCM, Mayo Clinic and MIMIC reports, only aortic valve regurgitation peak velocity was absent from the Northwestern dataset. In addition, mitral valve mean gradient was absent from the WCM echo reports and mitral valve orifice area was absent from the MIMIC and Mayo echo reports. At WCM, aortic valve mean gradient, aortic valve orifice area, inter-ventricular septum dimension at end diastole, left atrium size at end systole, left ventricular dimension at end diastole and left ventricular dimension at end systole were also found in high frequency. Among Mayo, Northwestern and MIMIC echo reports, left atrium size at end systole was found frequently. Both WCM, Northwestern and MIMIC reports showed high recall and precision in extracting

LVEF, with an F-score of 0.99, 0.99 and 0.95 respectively. We observed that *EchoExtractor* showed high precision (99%), while low recall (55%) for extracting LVEF from Mayo Clinic echo reports. Across all four datasets a F-score >0.90 was found on only two target concepts: aortic valve regurgitation and left atrium size at end systole. At WCM, Mayo Clinic and Northwestern, an F-score >0.90 was observed for aortic valve regurgitation, left atrium size at end systole, mitral valve regurgitation, and tricuspid valve regurgitation. Between WCM, Northwestern and MIMIC reports, a F-score >0.90 was observed for the concepts aortic valve regurgitation, aortic valve stenosis, left atrium size at end systole, left ventricular ejection fraction, and tricuspid valve regurgitation. Between WCM and Northwestern sites, mitral valve regurgitation showed an F-score >0.90.

Table 2: Performance evaluation of *EchoExtractor* NLP pipeline on echocardiograms originated at three medical centers (WCM, Mayo Clinic, Northwestern) and MIMIC

| Concept | WCM | | | MAYO | | | Northwestern | | | MIMIC | | |
|---|---|---|---|---|---|---|---|---|---|---|---|---|
| | Recall(%) | Precision (%) | F-score | Recall(%) | Precision (%) | F-score | Recall(%) | Precision (%) | F-score | Recall(%) | Precision (%) | F-score |
| aortic valve mean gradient | 100 | 100 | 100 | A | A | A | 39 | 100 | 56 | A | A | A |
| aortic valve orifice area | 100 | 77 | 87 | 25 | 75 | 38 | 8 | 64 | 13 | 0 | 0 | 0 |
| aortic valve regurgitation | 96 | 100 | 98 | 87 | 94 | 90 | 90 | 100 | 95 | 87 | 100 | 93 |
| aortic valve regurgitation peak velocity | A | A | A | A | A | A | A | A | A | A | A | A |
| aortic valve stenosis | 88 | 100 | 94 | 32 | 100 | 49 | 92 | 100 | 96 | 87 | 94 | 90 |
| e/e prime ratio | 0 | 0 | 0 | 25 | 50 | 33 | 94 | 100 | 97 | 100 | 100 | 100 |
| inter-ventricular septum dimension at end diastole | 99 | 100 | 100 | 38 | 25 | 30 | 27 | 98 | 43 | 100 | 38 | 56 |
| left atrium size at end systole | 98 | 100 | 99 | 98 | 96 | 97 | 92 | 99 | 96 | 85 | 98 | 91 |
| left ventricular dimension at end diastole | 100 | 100 | 100 | 0 | 0 | 0 | 28 | 100 | 44 | 0 | 0 | 0 |
| left ventricular dimension at end systole | 95 | 100 | 97 | 0 | 0 | 0 | 28 | 100 | 44 | A | A | A |
| left ventricular size | 67 | 97 | 79 | 73 | 94 | 82 | 39 | 100 | 56 | 38 | 99 | 54 |
| left ventricular ejection fraction | 99 | 99 | 99 | 55 | 99 | 71 | 97 | 100 | 99 | 91 | 100 | 95 |
| left ventricular posterior wall thickness at end diastole | 6 | 100 | 12 | 67 | 86 | 75 | 43 | 99 | 60 | 0 | 0 | 0 |
| mitral valve mean gradient | A | A | A | 6 | 100 | 11 | 39 | 100 | 56 | A | A | A |
| mitral valve orifice area | 100 | 6 | 11 | A | A | A | 31 | 100 | 48 | A | A | A |
| mitral valve regurgitation | 96 | 100 | 98 | 91 | 96 | 94 | 85 | 100 | 92 | 66 | 100 | 79 |
| mitral valve regurgitation peak velocity | A | A | A | A | A | A | 0 | | 0 | A | A | A |
| mitral valve stenosis | 60 | 100 | 75 | 17 | 100 | 29 | 83 | 100 | 91 | 40 | 100 | 57 |
| pulmonary artery pressure | 86 | 100 | 93 | 32 | 35 | 33 | 49 | 95 | 65 | 87 | 100 | 93 |
| right atrial pressure | 0 | 0 | 0 | 0 | 0 | 0 | 33 | 96 | 49 | 67 | 50 | 57 |
| tricuspid valve mean gradient | A | A | A | A | A | A | 100 | 7 | 13 | A | A | A |
| tricuspid valve orifice area | A | A | A | A | A | A | 100 | 100 | 100 | A | A | A |
| tricuspid valve regurgitation | 90 | 100 | 95 | 92 | 97 | 95 | 98 | 100 | 99 | 71 | 100 | 83 |
| tricuspid valve regurgitation peak velocity | 0 | 0 | 0 | 0 | 0 | 0 | 96 | 98 | 97 | 0 | 0 | 0 |

Error analysis of the echo reports from the study datasets revealed that false positive cases are mainly due to term mismapping or the reference range for a concept being wrongly mapped as the measured value. For instance, in WCM echo reports several target concepts are documented in semi-structured tabular format where a given concept is documented by the measured value followed by the reference value (e.g. "Aortic Valve Area   1.1 > 2.4 cm2"). When there are no measured value documented as in several instances of aortic valve orifice area (e.g. "Aortic Valve Area > 2.40 cm2") or left ventricular ejection fraction (e.g. "Ejection Fraction 0.55 - 0.75") or mitral valve orifice area (e.g. "Mitral Valve Area > 3 cm2"), then the NLP system extracted the value given for the reference range as the measured value.  For aortic valve orifice area and mitral valve orifice area, 23% and 94%, respectively, of reports were found to be false positive this way.  Term mismapping also leads to a higher proportion of false positives. For example, the term "e:a" ratio being wrongly mapped to e/e prime ratio as in "Mitral E:A Rt 2" accounting 18% of false positive cases for the target concept e/e prime ratio.  Although rare, the system wrongly mapped to terms "dilated aorta arch" to target concept left ventricular size with no apparent reason. In MIMIC echo reports, majority of false positive cases are due to term mismapping. Examples are "interatrial septum" being mapped to the concept interventricular septum dimension at end diastole, left ventricular cavity being mapped to the concept left ventricular dimension at end diastole, dilated descending aorta being mapped to left ventricular size, or mitral valve jet being mapped to tricuspid valve regurgitation peak velocity. Also when a short representation of a target concept is part of a word preceded by a space character; the system wrongly mapped that concept. An example is "ava" for aortic valve orifice area being mapped to the word "available". Also, some false positive cases were due to wrong mapping of the qualitative assessment of one concept to another. For example, the assessment "mildly thickened" of aortic valve leaflets was mapped to aortic stenosis as in  "The aortic valve leaflets (3) are mildly thickened but aortic stenosis is not present."

Error analysis on false negative cases revealed several instances where the system missed the target concept due to variations in the terms used to represent that concept in different clinical settings. For example, in WCM echo reports, the concept aortic valve max pressure gradient are consistently documented under peak aortic gradient, and since this is not one of the lookup terms for that concept and missed entirely. A second cause for false negative cases was when a line break (\n) or special characters (;, ~ , &), or parentheses appear between a concept and the corresponding value or qualitative assessment. Yet another cause for false negative cases was when a term for a target concept appears in two lines (e.g. right "\n" atrial pressure), and the system failed to extract the concept. Similarly, when two target concepts appeared too close to each other, the system failed to map those concepts with the correct value or assessment. For example, in MIMIC echocardiograms, there were several instances where aortic stenosis was described along with aortic valve orifice area (e.g. Mild AS (AoVA 1.2-1.9cm2)), and the system failed to correctly extract aortic valve area.

For the Mayo Clinic's echo reports, we found that semi-structured table format had negative impact on the performance of the concept extraction. For example, "e/e prime ratio" listed in the table entries could not be detected consistently and accurately. Therefore, the semi-structured tables were excluded from the performance evaluation. Otherwise, the Mayo Clinic's echo reports appeared to have relatively consistent syntax.  We also found that semicolon (;) was used in place of colon (:) for the representation of some concept values such as "Calculated left ventricular ejection fraction; 65 %", which resulted in roughly half of the false negatives being recorded for this concept.

In addition, some phrase patterns appeared to be more commonly used  at Mayo Clinic (e.g. Mitral valve sclerosis without stenosis), which resulted in 58% of the false negatives for the detection of the mitral valve stenosis concept and 47% of the false negatives for the aortic valve stenosis concept.  A similar issue occured for "Aortic valve systolic mean Doppler gradient 12 mmHg:, where the insertion of "Doppler" term caused a failure to detect the gradient concept.  The high occurrence rate of these specific phrase patterns appeared to be institution-specific and needed to be handled in the future.

At Northwestern, we also found multiple semi-structured table formats in the echocardiogram reports, which also led to poor NLP performance for some of the target concepts, Also we found many of the other sources of errors as in other datasets, including the interjection of spacing and special characters in the reports, omissions of spacing between concepts and values (e.g., Ao valve open2.2 cm), different concept terms used, and multiple concepts in the same sentence.  Of the few false positives we found, the NLP did not recognize some less common abbreviations such as those for aortic insufficiency (AI) that were not listed in most reports. Many of the false negatives were due to the above reasons; in particular, multiple concepts were frequently missed as they were in a separate table and the description of the measures were partly abbreviated (vs. fully spelled out or completely abbreviated as in other reports). For example,  left ventricular size measures such as left ventricular dimension at end systole were listed in the table as "LV Size-end systole" instead of "LVESD" or "left ventricular size-end systole." The remaining few false negatives were due to missing values in sentences where other concepts were mentioned. For example, some reports started their summary section at the end of the report with summarizing multiple normal measures in 1 statement, such as "Left ventricular size, systolic function, wall thickness, and wall motion are all normal." There were very few instances of FNs where the value extracted by the NLP was the wrong value for the concept-value pair; thus, the NLP algorithm was good at finding the correct value when it found a concept.

**Discussion**

Measurements on cardiac structure and function that can be derived semi-automatically from echocardiogram reports can enable clinical and outcomes research. Echocardiography provides detailed information regarding cardiac structure and function in an easily accessible and cost-effective manner[23]. Concepts such as wall thickness and left ventricular dilation is associated with cardiovascular disease and heart failure[24,25], left atrial size is related to incidence of atrial fibrillation, stroke, and death, and aortic root size is associated with risk of heart failure, stroke, and mortality[26]. While many modern EHR systems record various concepts from echocardiograms in a more structured manner, semi-structured or unstructured echocardiogram reports are still the primary source in most clinical settings. In the past a number of NLP methods have been developed to extract various cardiac measures documented in echocardiography reports. While majority of these studies focused on the extraction of LVEF[12,27-29], some have attempted to extract a broader range of cardiac concepts[21,30]. However, to the best of our knowledge these methods have not been tested for their performance on data originated from clinical sites other than the original NLP algorithm and method development site. As a result, "true portability" of the developed NLP systems has never been rigorously assessed. The present study fill this knowledge gap by adopting a system developed initially by the VA, and then subsequently implemented at three non-VA medical centers without any modifications, to test its performance in processing echocardiogram reports.

In particular, the NLP system under investigation - *EchoExtractor* - was designed to extract 27 concepts from narrative text, echocardiograms, and radiology reports, and the original developers reported an average precision of concept-value extraction of 0.982 in echo documents, 0.969 in Radiology, and 0.936 in narrative text, with F-score of 0.844, 0.877, and 0.872 respectively. Specifically, the system reported precision and recall values for 20 data elements in echocardiograms. The remaining 7 target concepts were either absent or too low in total mentions to report performance concepts. Out of the 20 target concepts in echocardiograms, 17 elements have been reported with a precision of 0.90 or higher[21].

When compared to the reported data, the overall performance of the NLP system was found to be lower on all three study sites. In WCM dataset, target concepts such as inter-ventricular septum dimension at end diastole, left ventricular dimension at end diastole, and left ventricular dimension at end systole are documented as part of a tabular format within echo reports and appeared consistently across all reports, and hence resulted high performance for these concepts. On the other hand, concepts such as aortic valve orifice area and mitral valve orifice area, although appear in a tabular format and are typically documented along with a reference range in the echo reports. For these concepts we observed that the NLP system often failed to correctly identify the measured values from the reference range, thereby resulting in lower precision.

The Mayo Clinic dataset comprised of semi-structured tables to represent the echocardiography concepts. While the table formats had modest variations over the years, the NLP system was generally unable to detect concepts located in these tables successfully. Our evaluations suggest that the representation of target concepts using a table format is likely to decrease the NLP system performance, and alternative approaches to capturing data in these sections should be evaluated. Specifically, we observed 16 target concepts in the Mayo Clinic dataset, out of which the NLP system achieve a high precision and F-score >0.90 for only 4 concepts. A notable exception in Mayo Clinic dataset compared to other datasets is the low recall observed on LVEF. While WCM, Northwestern and MIMIC echo reports showed high recall (99%, 97%, and 91%, respectively), Mayo Clinic echo reports showed only 55 percent. On further analysis, we concluded that the high rate of failure to detect LVEF is due to a semicolon ';' character (e.g. ejection fraction; 67) that appeared between the concept and value, which was causing system error. Common institutional phrasings such as "sclerosis without stenosis" that do not fit into the predefined patterns can be a significant source for concern and poor performance.

We observed 23 target measurements on Northwestern echo reports, out of which the NLP system showed a precision and F-score values >0.90 for 10 elements. Unlike echo reports from other sites, the concept tricuspid valve regurgitation peak velocity was found with high recall and precisions. Although this concept was mentioned infrequently in reports from other datasets, they were mentioned in ~50% reports at Northwestern using system recognized concept names. Similarly, the measurement tricuspid valve orifice area is found only in Northwestern echo reports, and extracted with a high recall and precision.

Unlike the echocardiograms from WCM, Northwestern and Mayo, MIMIC reports are entirely in a free-text format. We observed 16 target concepts in the MIMIC dataset, out of which the NLP system achieved a high precision and F-score values for only 6 concepts. This relatively low performance on MIMIC dataset compared with the performance on other datasets, is probably due to the highly unstructured nature of these reports and variability in the lexical terms.

**Conclusion**

The work described in this paper details a case study to adopt a specific instance (*EchoExtractor*) of a NLP system (Leo) previously developed by the VA. The adopters (WCM, Mayo Clinic and Northwestern) individually led the NLP system implementation without any modification, drawing on existing resources, and employing conventional software skills. The study demonstrates that concept-value extraction from echocardiograms can vary depending on local text formats and variations in lexical terms used to document various concept across different clinical settings. The NLP system used a custom lookup dictionary as part of its extraction algorithm[21], and our analysis showed that the performance of the system for several of the target concept can be improved significantly by extending this lookup dictionary with additional terms specific to individual sites. Many of the false negative cases resulted because of the absence of terms in the custom dictionary. Despite the fact that this system was developed based on documents originated at thousands of points of care and authored by a multitude of clinical professionals within the VA, it also suffers from some of the same drawbacks as other NLP systems have reported including the need for a local customization of extraction logic and extending the lookup dictionary based on local data source.

**Acknowledgements**

This work was supported in part by NIGMS grant R01GM105688, NHGRI grants U01HG008673 and U01HG6379, NCATS grant UL1TR001422. The content is solely the responsibility of the authors and does not necessarily represent the official views of the National Institutes of Health.